\pdfoutput=1 

\documentclass[conference]{IEEEtran}

\usepackage[T1]{fontenc} 
\usepackage[utf8]{inputenc} 
\usepackage[english]{babel} 

\usepackage[
    backend=biber, 
    style=ieee, 
    hyperref, 
    maxbibnames=99, 
    doi=false, 
    isbn=false, 
    ]
        {biblatex} 
\usepackage[
    autostyle] 
        {csquotes} 

\usepackage[top=1in, bottom=0.75in, left=0.75in, right=0.75in]{geometry}
\usepackage{booktabs}
\usepackage[pdftex]{graphicx}
\graphicspath{{figures/}}
\DeclareGraphicsExtensions{.pdf,.jpg,.png}
\usepackage[cmex10]{amsmath}
\usepackage{amssymb}
\usepackage{bm} 
\usepackage[font=footnotesize,caption=false]{subfig} 
\usepackage{booktabs}
\usepackage{interval}
\usepackage{tabularx}
\usepackage{url}
\usepackage{hyperref}

\DeclareMathOperator{\parents}{parents}
\newcommand{\Effect}{\ensuremath{\text{Effect}}}
\newcommand{\EffectX}{\ensuremath{\text{EffectX}}}
\newcommand{\EffectY}{\ensuremath{\text{EffectY}}}

\begin{document}

\title{Learning at the Ends: From Hand to Tool Affordances in Humanoid Robots}

\author{\IEEEauthorblockN{Giovanni Saponaro\IEEEauthorrefmark{1},
Pedro Vicente\IEEEauthorrefmark{1},
Atabak Dehban\IEEEauthorrefmark{1}\IEEEauthorrefmark{2},
Lorenzo Jamone\IEEEauthorrefmark{3}\IEEEauthorrefmark{1}, \\
Alexandre Bernardino\IEEEauthorrefmark{1} and
José Santos-Victor\IEEEauthorrefmark{1}}
\IEEEauthorblockA{\IEEEauthorrefmark{1}Institute for Systems and Robotics,
Instituto Superior Técnico,
Universidade de Lisboa, Lisbon, Portugal\\
Email: \{gsaponaro,pvicente,adehban,alex,jasv\}@isr.tecnico.ulisboa.pt}
\IEEEauthorblockA{\IEEEauthorrefmark{2}Champalimaud Centre for the Unknown, Lisbon, Portugal}
\IEEEauthorblockA{\IEEEauthorrefmark{3}ARQ~(Advanced Robotics at Queen Mary), \\
School of Electronic Engineering and Computer Science,
Queen Mary University of London, United Kingdom \\
Email: l.jamone@qmul.ac.uk}}

\maketitle

\begin{abstract}
One of the open challenges in designing robots that operate successfully in the unpredictable human environment is how to make them able to predict what actions they can perform on objects, and what their effects will be, i.e., the ability to perceive object affordances. Since modeling all the possible world interactions is unfeasible, learning from experience is required, posing the challenge of collecting a large amount of experiences~(i.e., training data).
Typically, a manipulative robot operates on external objects by using its own hands~(or similar end-effectors), but in some cases the use of tools may be desirable; nevertheless, it is reasonable to assume that while a robot can collect many sensorimotor experiences using its own hands, this cannot happen for all possible human-made tools.

Therefore, in this paper we investigate the developmental transition from hand to tool affordances: what sensorimotor skills that a robot has acquired with its bare hands can be employed for tool use?
By employing a visual and motor imagination mechanism to represent different hand postures compactly, we propose a probabilistic model to learn hand affordances, and we show how this model can generalize to estimate the affordances of previously unseen tools, ultimately supporting planning, decision-making and tool selection tasks in humanoid robots. We present experimental results with the iCub humanoid robot, and we publicly release the collected sensorimotor data in the form of a hand posture affordances dataset.
\end{abstract}


\IEEEpeerreviewmaketitle

\section{Introduction}

Robotics is shifting from the domain of specialized industrial manipulators operating in a predictable and repeatable environment, towards more adaptable systems that are expected to work among people. This new generation of robots needs to operate in an unstructured and dynamic world, exposed to a multitude of objects and situations that cannot be modeled \emph{a~priori}. A crucial ability needed by these robots to succeed in such environment is to be able to predict the effects of their own actions, or to give a reasonable estimate when they interact with objects that were never seen before; this applies both to objects that are acted upon, and to those that may be used as tools to extend the capabilities of the robot body. The interplay between these elements (actions, objects, tool, effects) can be described by resorting to the concept of affordances~(action possibilities), a concept that was initially proposed in psychology~\cite{jgibson:2014}, supported by neuroscience evidence, and applied to several artificial intelligence~(AI) and robotics systems; indeed, a growing line of research takes inspiration from human intelligence and development in order to design robust robot algorithms for affordance learning and perception, that can, in turn, support action selection and planning in unstructured environments~\cite{jamone:2016:tcds}.

\begin{figure}
    \centering
    \includegraphics[width=0.9\linewidth]{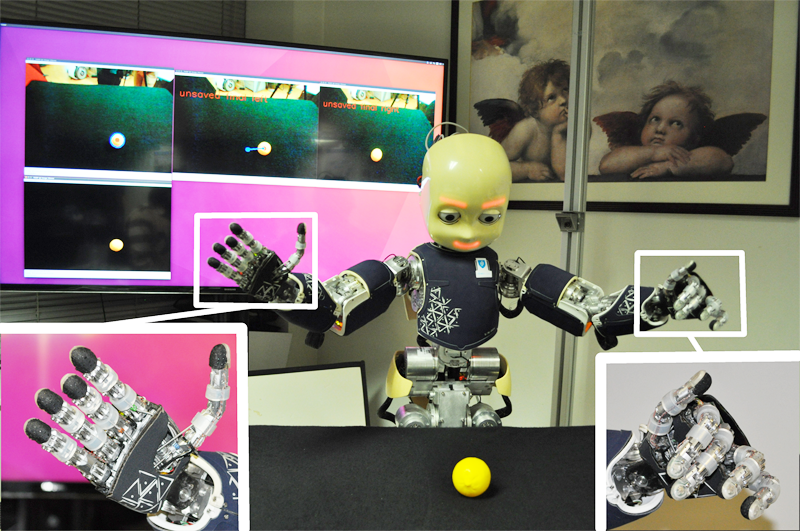}
    \caption{The iCub humanoid robot performing motor actions with different hand postures onto a physical object. In the background screen, we show the visual routines that monitor the evolution of the environment.}
    \label{fig:cover}
\end{figure}

In humans, learning the affordances of one's own hands~(i.e., what actions one can do with them, what effects one can obtain) is a long developmental process that begins in infancy~\cite{egibson:1994,james:2010:icd}, and continues during childhood through exploration of different actions and different objects~\cite{rosenblatt:1977:bioplay,rosenbaum:2009}. Functional tool use appears later~\cite{lockman:2000:childdev,szokolszky:2010,lobo:2013:ibd,fagard:2014:emergence}, and the knowledge previously acquired by babies during manual exploration of objects is likely to play a role. Definitely, one of these roles is that the increased hand dexterity acquired during development allows the child to correctly grasp, manipulate and orient a tool; however, another role may be that the child ``sees'' in the shapes of some tools relevant characteristics that remind the child of previously used shapes of the own hands~(although no experimental evidence of this perceptual skill has been provided in the developmental psychology literature, as far as the authors know).

Inspired by these observations in developmental psychology, and motivated by a clear need of autonomous robotic systems, in this paper we investigate a possible developmental \emph{transition from hand to tool affordances}. In particular, we explore how a learned representation of hand affordances can be generalized to estimate the affordances of tools which were never seen before by the robot. We train a probabilistic model of affordances, relating visual features of (i)~different robotic hand postures and (ii)~different objects, with the resulting effects caused by the robot motor actions onto such objects; training data are collected during several experiments in which the iCub humanoid robot performs manual actions on objects located on a table (see Fig. \ref{fig:cover}).
Our probabilistic model is implemented as a Bayesian Network~\cite{murphy:2012:mlprob}, we publicly release a novel dataset of hand posture affordances~(\url{http://vislab.isr.tecnico.ulisboa.pt/} $\rightarrow$ Datasets), and we test it for generalization against an available dataset of tool affordances~\cite{dehban:2016:eccvws}.

This article is structured as follows. Sec.~\ref{sec:related_work} overviews related work about hand and tool affordances in the contexts of psychology and robotics. Sec.~\ref{sec:proposed_approach} explains our proposed framework for learning the affordances of different hand postures on a humanoid robot, and how its generalization ability can prove fruitful for robot autonomy. Sec.~\ref{sec:results} reports our current experimental results. Finally, in Sec.~\ref{sec:conclusions} we give concluding remarks and discuss the significance and scope of our contribution.

\section{Related Work}
\label{sec:related_work}

Several researchers have investigated the role of hand actions during human intelligence development for learning to deal with the uncertainty of the real world~(e.g., toddler visual attention~\cite{yu:2009:tamd}) and \emph{tool use}.
Piaget documents an observation where his daughter makes an analogy between a doll's foot hooking her dress, and her own finger bent like a hook~\cite{piaget:1962}.
Tool use competence in humans emerges from explorative actions, such as those performed with the child's bare hands in the first year~\cite{smith:2014:jecp}.
A longitudinal study on infants~\cite{fagard:2014:emergence} shows that, at around 16--20 months, children start to intentionally and successfully bring faraway toys closer to themselves with the help of tools such as rakes.

Lockman~\cite{lockman:2000:childdev} suggests that the actions employed by toddlers on a daily basis with their everyday objects in their surroundings, likely put forward the idea that these actions initially incorporate many of the~(previously learned) motor patterns that infants employ with their hands and arms for exploring and learning. Szokolszky~\cite{szokolszky:2010} stresses how tool use is dependent and continuous with other percepts: action routines, such as reaching, grasping, focusing on an object or on a person, and eating with the hand.
In~\cite{lobo:2013:ibd}, Lobo highlights the following points about the relationship between early self-exploration behaviors and developing object exploration behaviors: (i)~infants are already actively engaging in exploratory behaviors to inform themselves about the affordances of their own bodies, objects, and the intersect of the two in the first months of life; (ii)~the emergence of reaching is an important step forward towards advanced object exploration and advanced self-exploration; (iii)~the behaviors that infants use to explore their own bodies and surfaces during the first months of life may form the set of behaviors from which they later \emph{choose}, as they begin to interact with objects.
It is with these considerations in mind that, in this paper, we pursue a robotic model that transfers hand knowledge to tool knowledge, and in the experimental part we verify the applicability of our model for task-dependent \emph{tool selection}, given previous exploratory hand knowledge.

In robotics, there is an increasingly large body of literature in affordances and computational models of affordances. A recent overview is given in~\cite{jamone:2016:tcds}. The idea behind robots learning object affordances, or action possibilities, is that knowledge acquired through self-exploration and interaction with the world can be used to make inferences in new scenarios and tasks, such as prediction~\cite{montesano:2008}, tool use~\cite{stoytchev:2008:lnai,tikhanoff:2013:humanoids,goncalves:2014:icarsc,goncalves:2014:icdl}, and planning~\cite{kruger:2011:ras,antunes:2016:icra}.
Recently, Schoeler and Wörgötter~\cite{schoeler:2016:tcds} introduced a framework for analyzing tools by modeling their dynamics and providing object graphs based on their parts,
to support the conjecture of a cognitive hand-to-tool transfer~(e.g., to understand that a helmet and a hollow skull can both be used to transport water).
However, to the best of our knowledge, our work is the first contribution, in the robot affordances field, which explicitly looks at the visuomotor possibilities offered by different hand morphologies and postures~(e.g., hands with extended fingers, hands with closed fingers), and exploits this information to acquire~(through self-exploration) a model that is able to generalize to novel situations for a robotic agent, including making the crucial developmental leap from hand use to tool use, as observed in babies by psychology studies.

\section{Proposed Approach}
\label{sec:proposed_approach}

\begin{figure*}
    \subfloat
    {\includegraphics[width=0.3\linewidth]{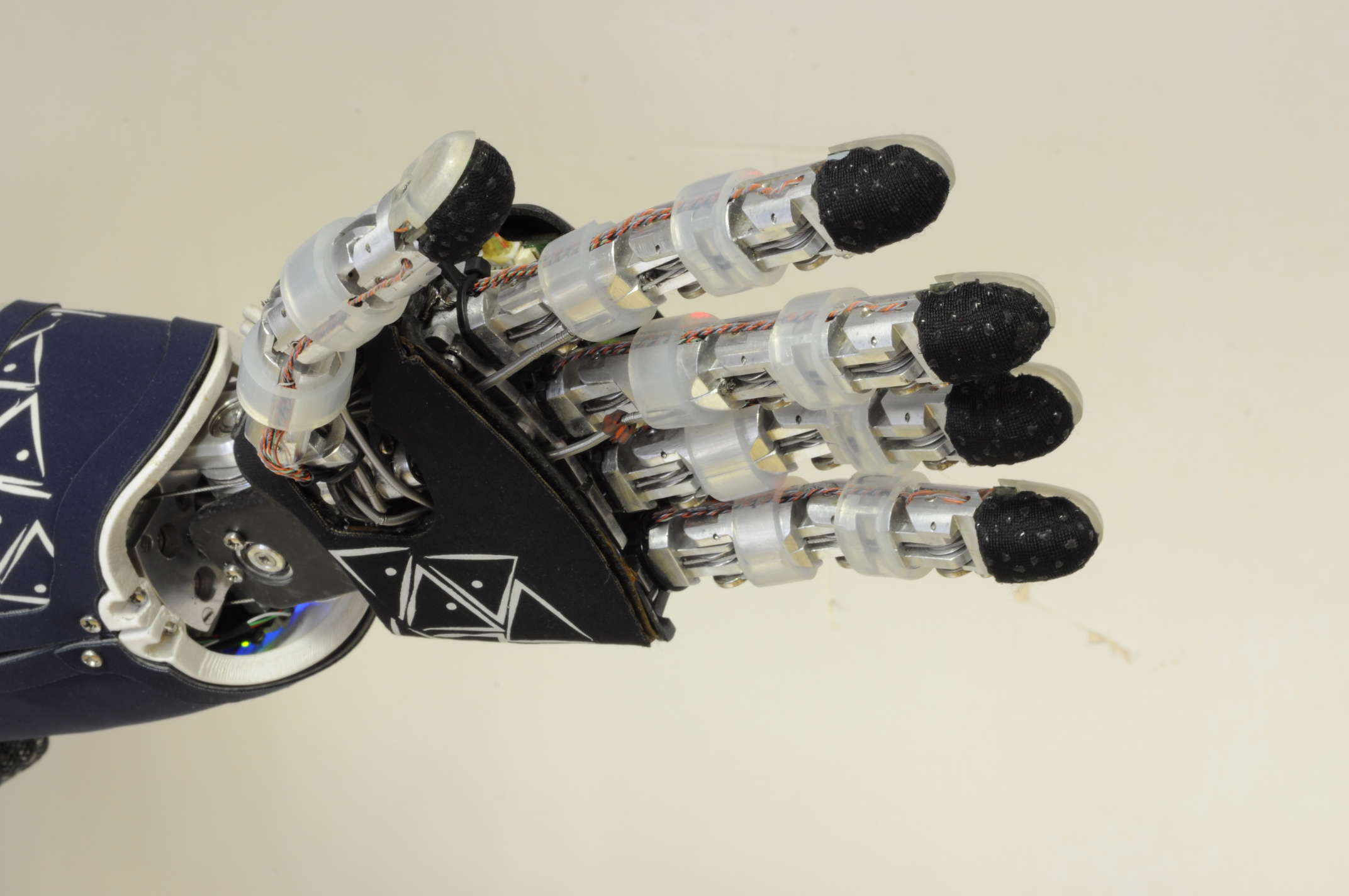} } \quad
    \subfloat
    {\includegraphics[width=0.3\linewidth]{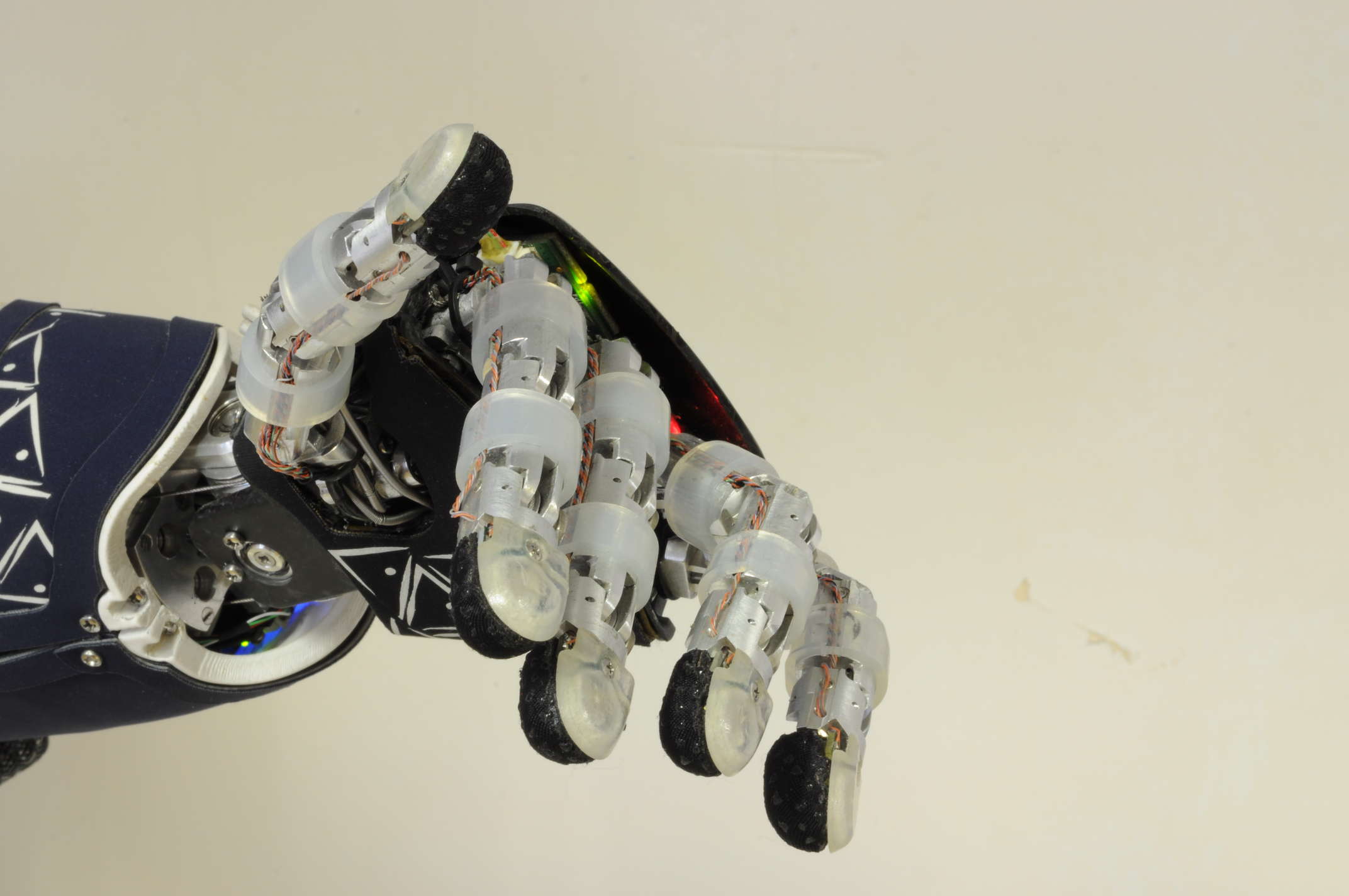} } \quad
    \subfloat
    {\includegraphics[width=0.3\linewidth]{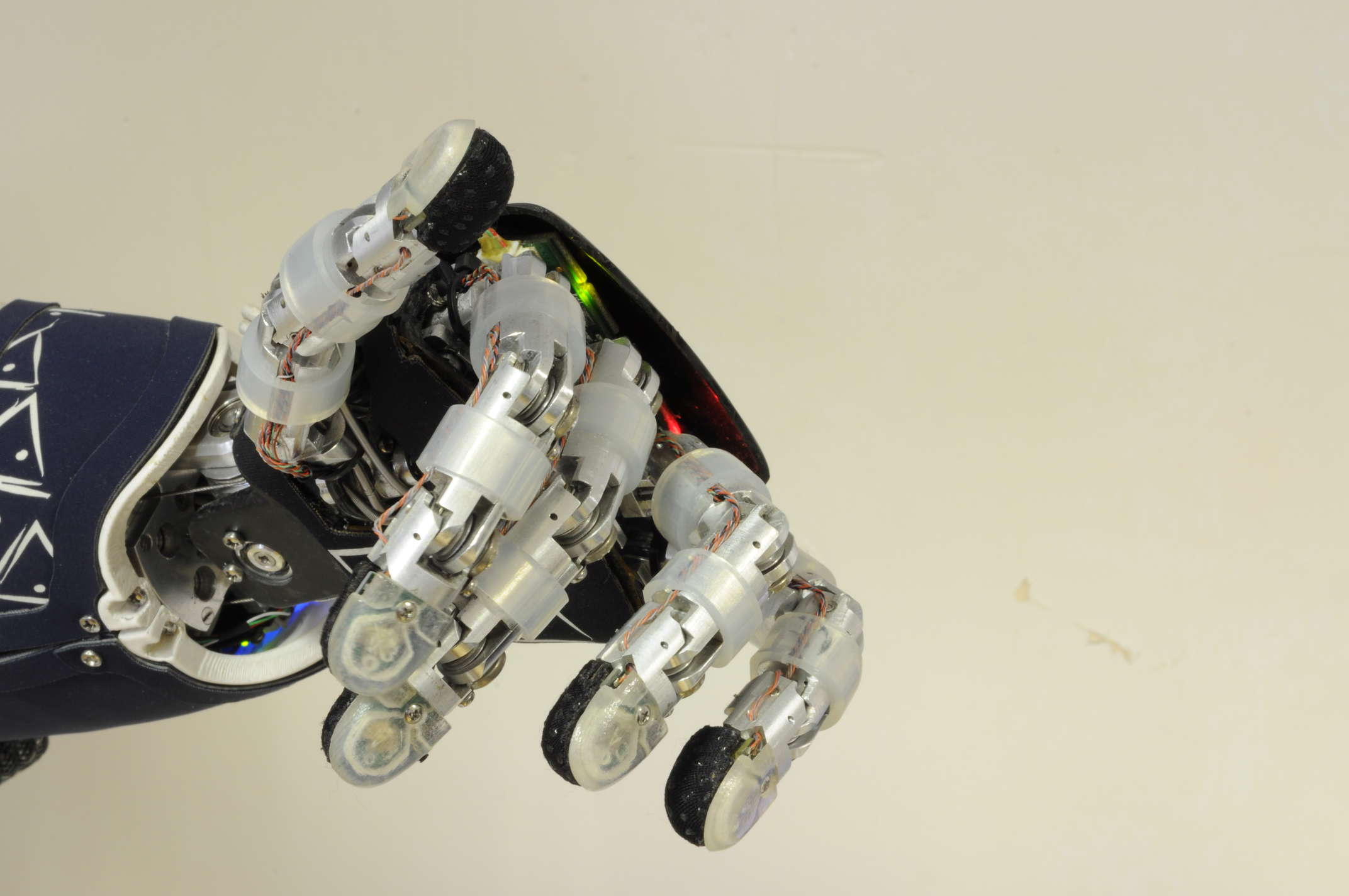} } \\
    \subfloat
    {\includegraphics[width=0.3\linewidth]{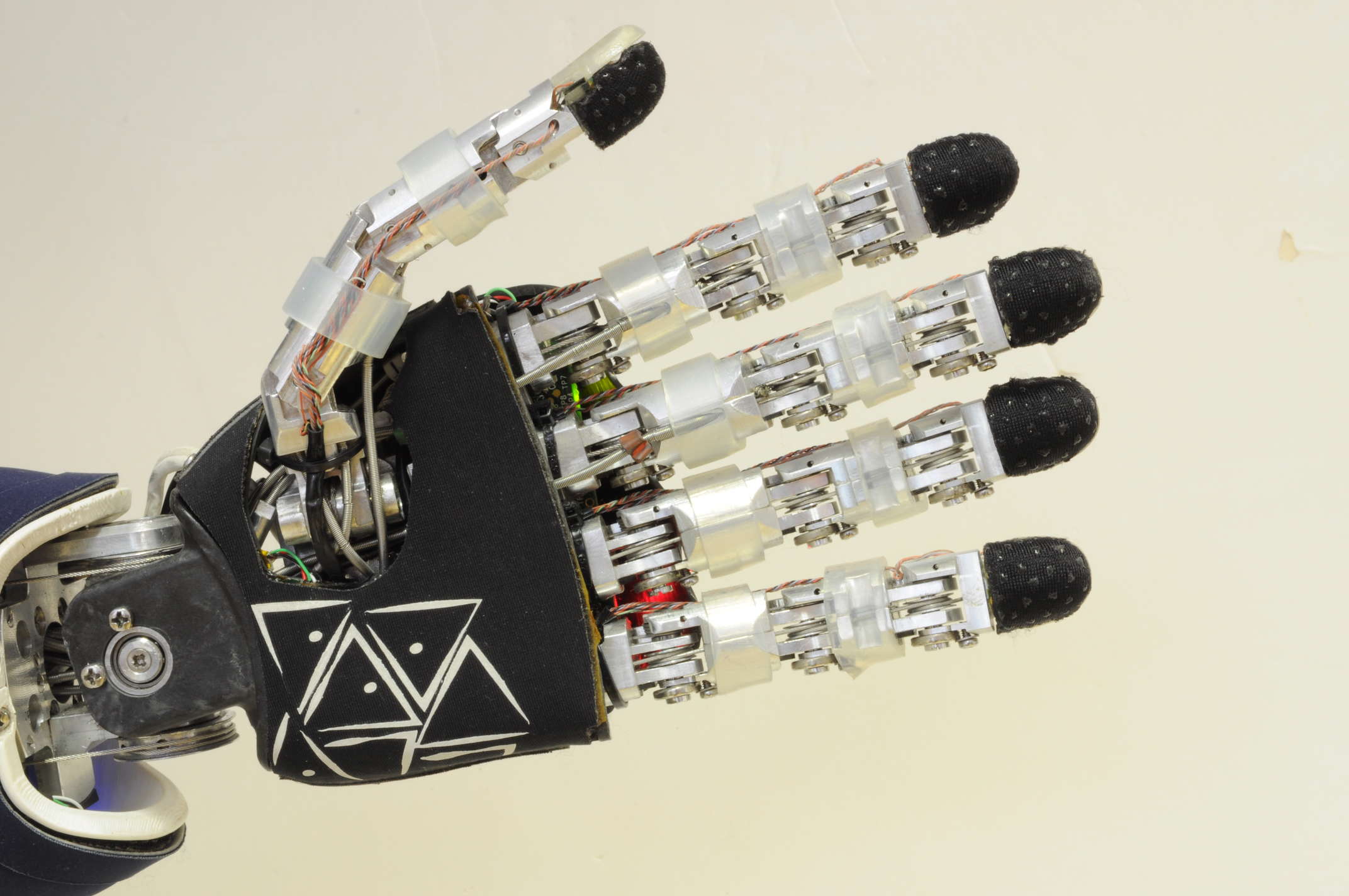}  } \quad
    \subfloat
    {\includegraphics[width=0.3\linewidth]{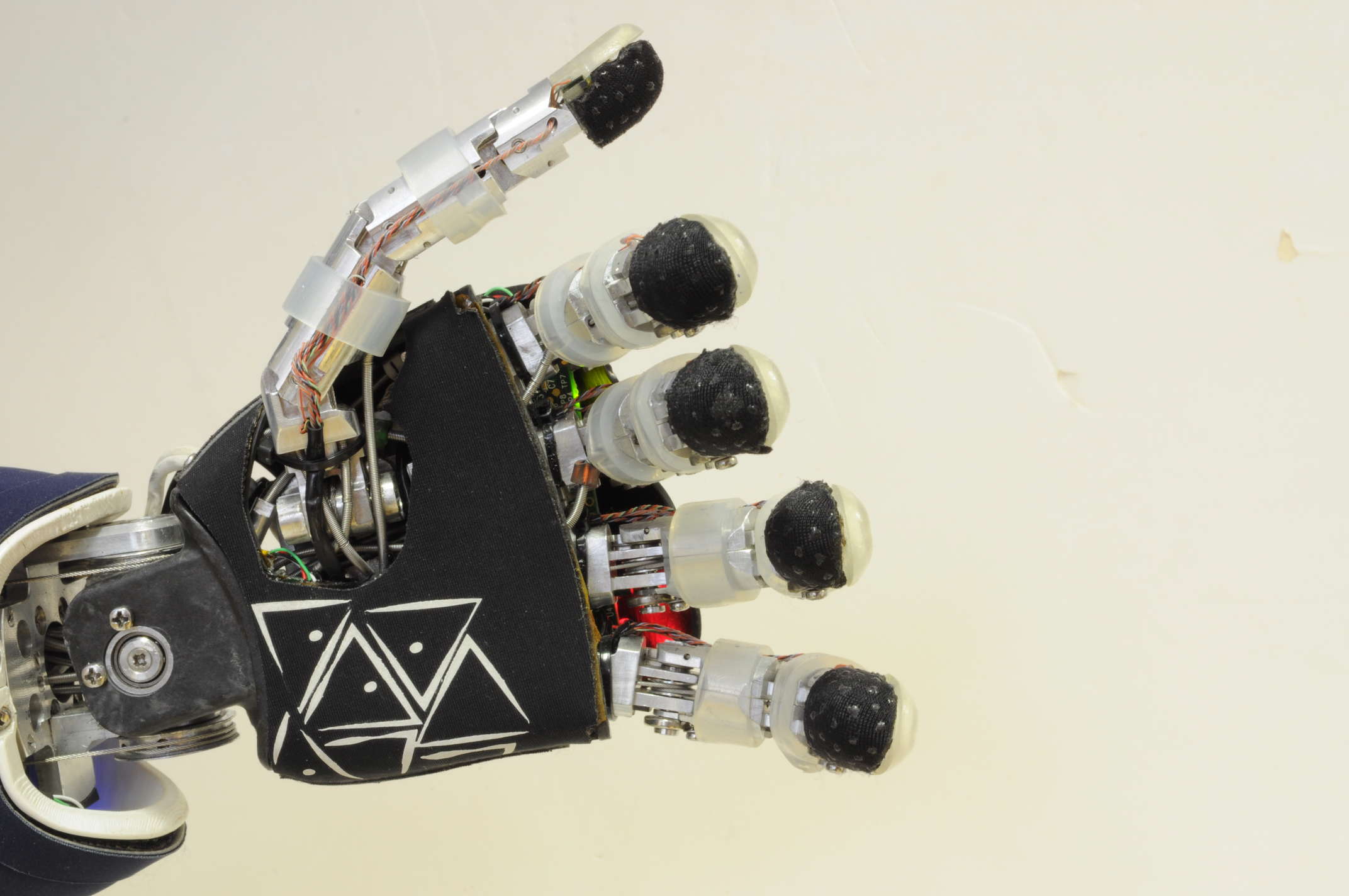} } \quad
    \subfloat
    {\includegraphics[width=0.3\linewidth]{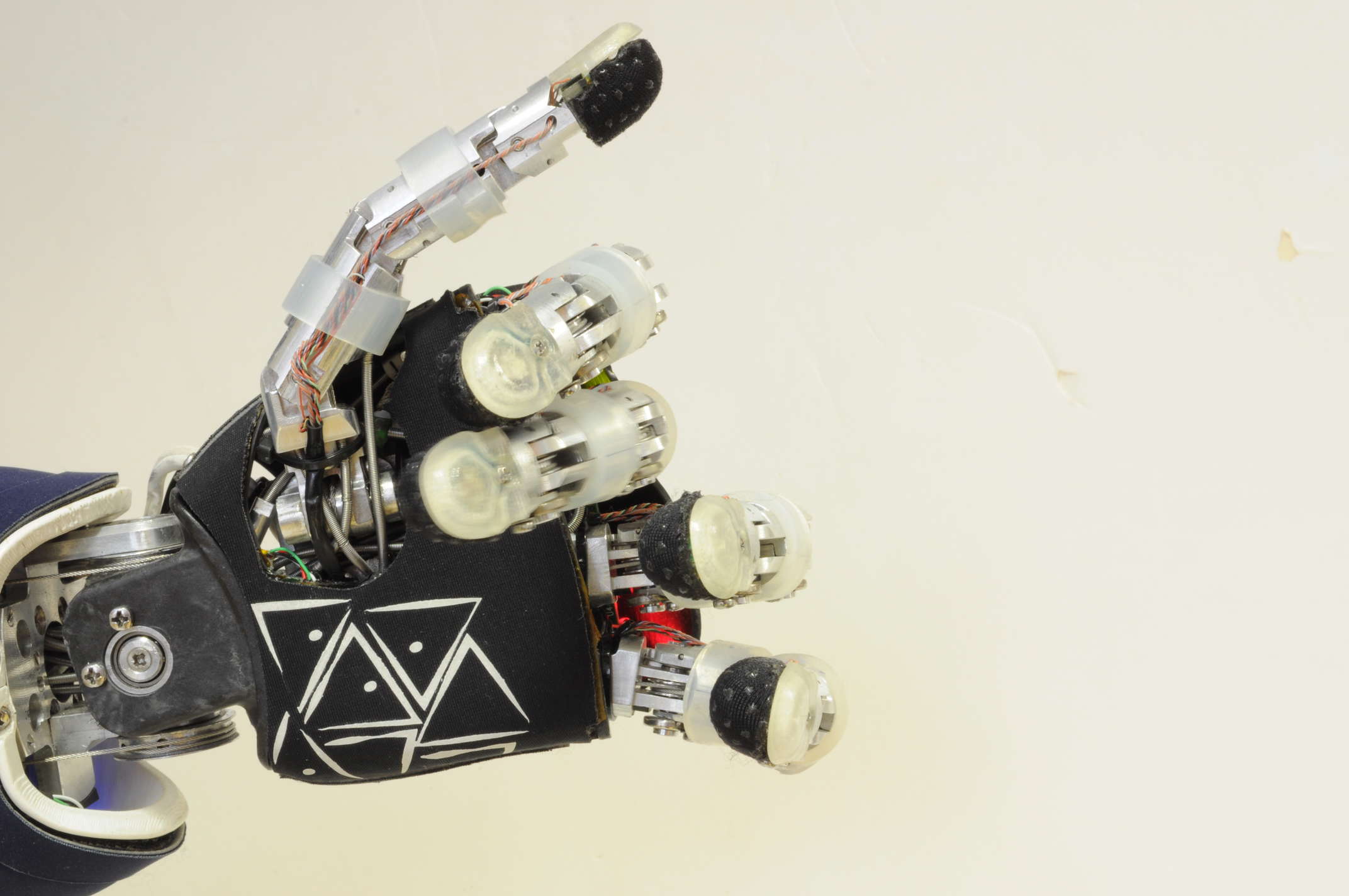} } \\
    \subfloat
    {\includegraphics[width=0.3\linewidth]{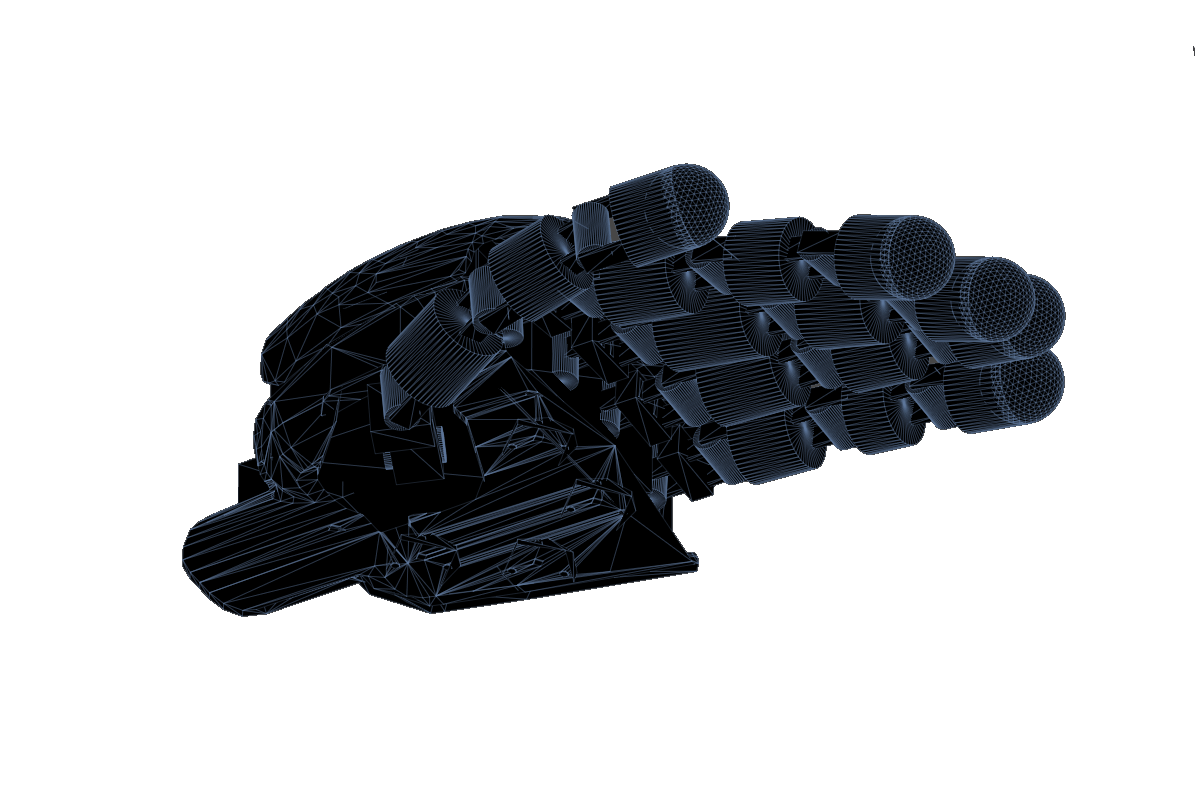} } \quad
    \subfloat
    {\includegraphics[width=0.3\linewidth]{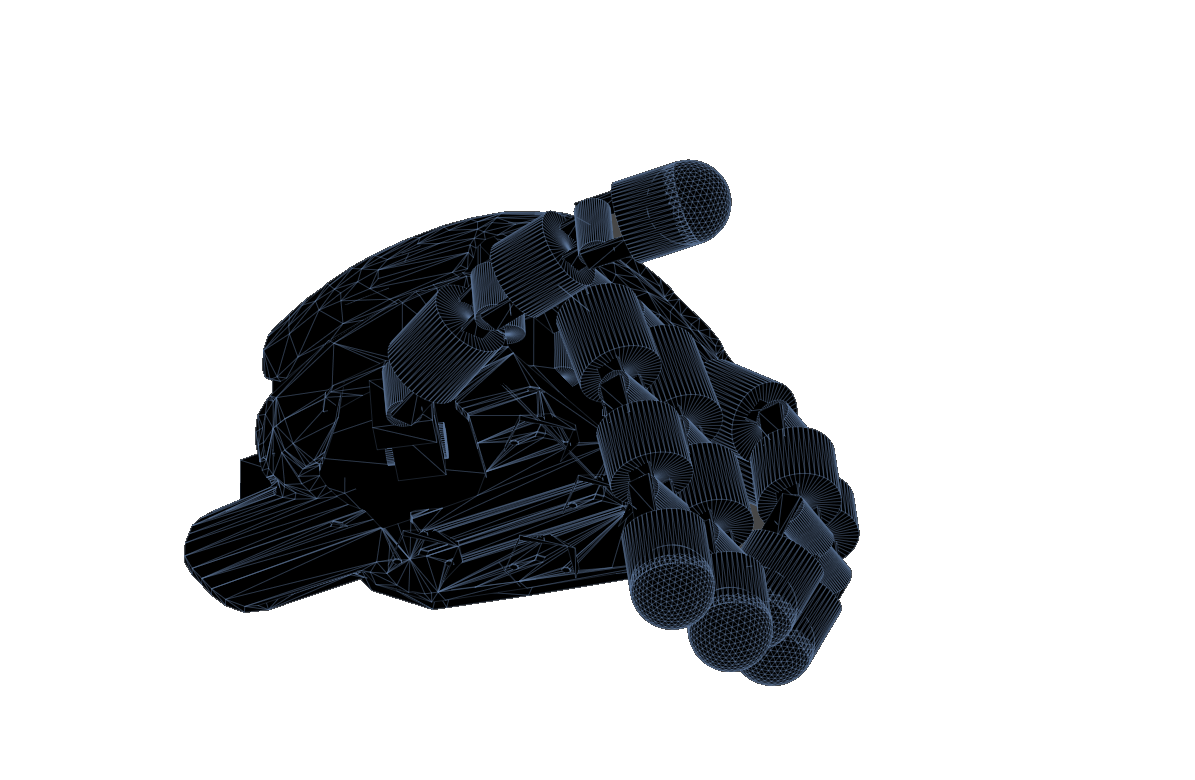} } \quad
    \subfloat
    {\includegraphics[width=0.3\linewidth]{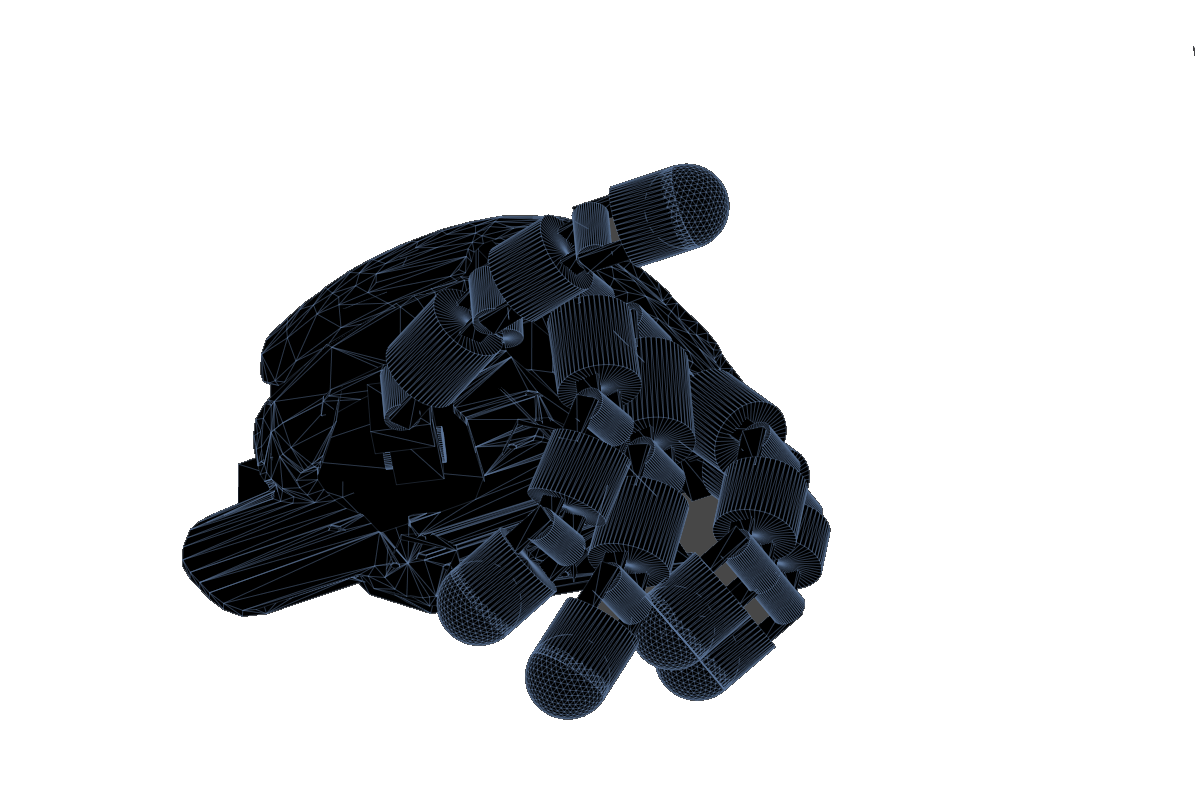}}
    \caption{The three hand postures adopted in this work. Left column: straight hand; center column: bent hand; right column: arched hand. The first two rows are real robot postures seen from different viewpoints; the last row shows the simulated body schema CAD model of the top viewpoint. From the latter simulated view, we obtain the segmented silhouette contour of the hand and its shape features (see Sec.~\ref{sec:descriptors}).}
    \label{fig:postures}
\end{figure*}

In this paper, we investigate how a humanoid robot can learn the affordances of different hand postures, similarly to the developmental processes observed during the first year of life in babies. We use a probabilistic model of affordances which relates (i)~visual features of the agent's own hands, (ii)~visual features of a target object located on a surface, (iii)~a motor action, and (iv)~the resulting effects of the action onto the object, in the sense of the physical displacement compared to the initial position. We currently use three different robot hand postures, shown in Fig.~\ref{fig:postures}.

\subsection{Motor Control}
\label{sec:motor_control}

The motor actions that we provide to the robot for learning the affordances of its actions performed with bare hands are: tapping an object from the left side~(with the palm of the hand), tapping an object from the right side~(with the back of the hand), pushing an object away from the agent, and drawing the object towards the agent~(i.e., pulling). The low-level control routines to realize the motor behaviors are based on works and software modules previously made available by other researchers in the iCub community~\cite{pattacini:2010:iros,roncone:2016:rss}. The location of the target object (i.e., the location where the robot performs an action) can be anywhere on the table, provided that it is within the reachable space of the robot end-effector, and that it satisfies geometric safety limits to avoid self-collisions. We determine this location with visual segmentation routines.

\subsection{Internal Model and Hand Imagination}
\label{sec:hand_imagination}

From a developmental psychology perspective, body awareness appears to be an incremental learning process that starts in early infancy~\cite{vonhofsten:2004:tcs} or probably even prenatally~\cite{joseph:2000:drev}. Such awareness is supported by a neural representation of the body that is constantly updated with multimodal sensorimotor information acquired during motor experience and that can be used to infer the limbs' position in space and guide motor behaviors: a body schema~\cite{berlucchi:1997:tneu}. The body schema is not necessarily restricted to proprioceptive and somatosensory perception \textit{per se}, but can also integrate visual and perhaps auditory information. Moreover, sub-representations of the body according to their function are also proposed by several authors. The visual information is usually called body image~\cite{riva:2014:fnhum}.

In this work, the learning of the sensorimotor mapping of affordances is based on the joint-space representation. The body schema and the forward model are then exploited to simulate motor behaviors and imagining sensory outcomes. The same approach seems to characterize humans~\cite{miall:1996:nn}.
The hand imagination is implemented through an internal model simulator developed within a previous work of ours~\cite{vicente:2016:jint}. From a technical perspective, using the simulated robot rather than the real one to obtain the hand posture visual shape, serves to filter out noise from the image processing pipeline. Although it is not always true that we can generalize from simulation to the real robots, in this case, we adopt a graphically and geometrically precise appearance model of the robotic hand~(based on the CAD model), therefore we can use the internal model simulation without losing generality or compromising the overall idea (see the real and simulated hands in Fig.~\ref{fig:postures}).

\subsection{Visual Features}
\label{sec:descriptors}

We characterize hand postures and physical objects by their shape~(i.e., the contour or silhouette obtained with segmentation algorithms), and we extract exemplary 2D visual features from them.
The features that we extract are pre-categorical shape descriptors computed as geometric relationships between perimeter, area, convex hull and approximated shapes of the segmented silhouettes of the objects in front of the robot, and they are based on~\cite{zhang:2004:shape}. Our visual feature extraction is similar to our previous works~\cite{goncalves:2014:icarsc,goncalves:2014:icdl,antunes:2016:icra}, however it now incorporates a richer set of~13 features instead of~5: convexity, eccentricity, compactness, circularity, squareness, number of convexity defects~(i.e., number of cavities along the contour, for example the ``holes'' between fingers in a hand image), and seven central normalized moments. It is worth noting that we do not classify or label our shapes into classes, but we merely reason about their shape-derived features: this gives our system some flexibility, making it able to process unknown situations not seen during training.

\subsection{Affordance Learning}

\begin{figure}
    \centering
    \includegraphics[width=0.9\linewidth]{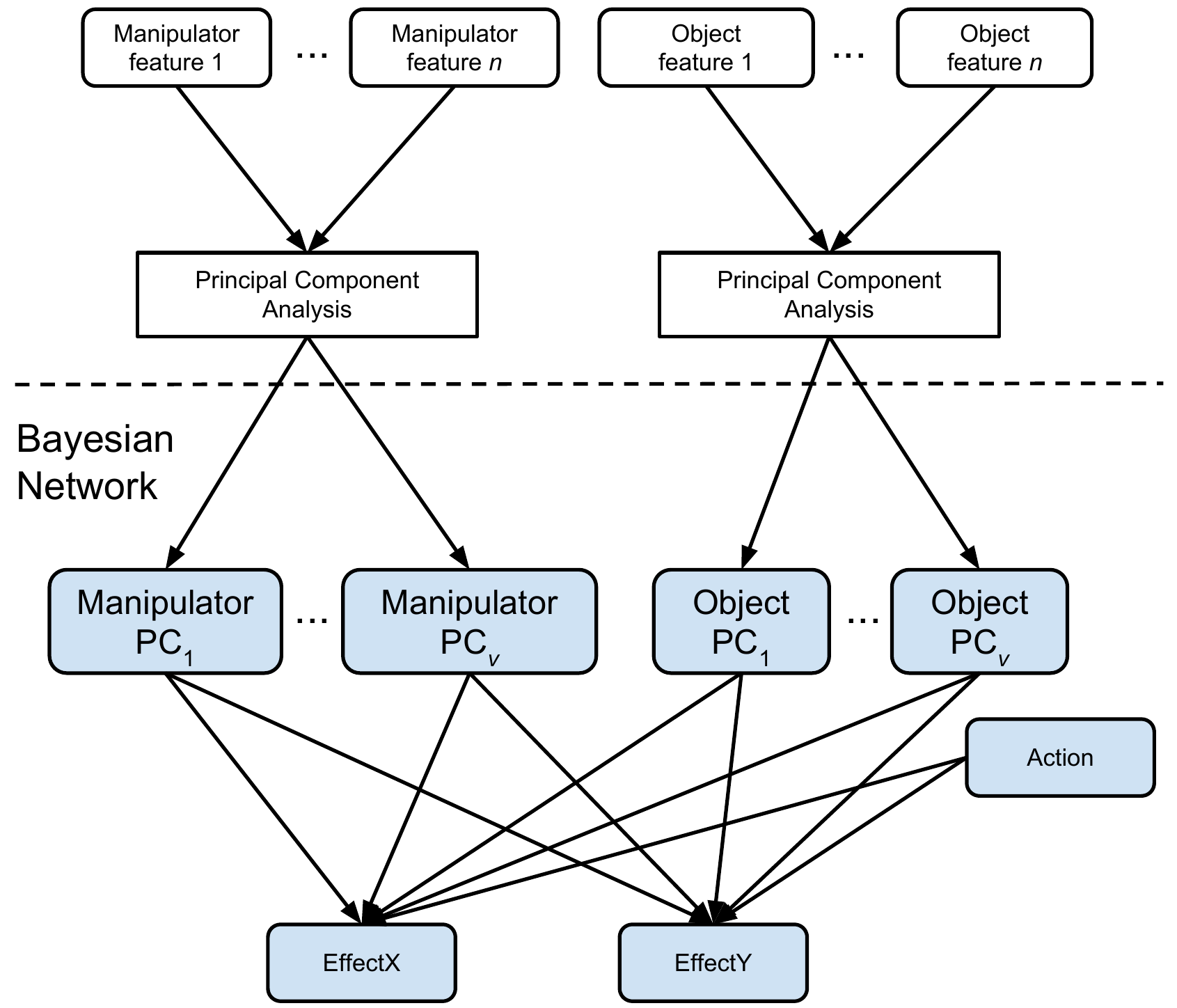}
    \caption{Structure of the Bayesian Network that we use to model the affordance relationships between robot manipulator, target object, motor action, and resulting effects, similarly to~\cite{goncalves:2014:icdl}.
    The raw visual features of manipulators and objects are real-valued and normalized between~0 and~1~(see Sec.~\ref{sec:descriptors}).
    In the network, Principal Components of manipulators and objects are discretized into~2 bins, the action node is a discrete index~(see Sec.~\ref{sec:motor_control}), and the effect nodes are discretized into~5 bins, according to the hyper-parameters used during training, listed in Table~\ref{tab:pca_hyperparams}.}
    \label{fig:pca_bayesian_network}
\end{figure}

Fig.~\ref{fig:pca_bayesian_network} shows the structure of the Bayesian Network~\cite{murphy:2012:mlprob} that we train with robot self-exploration hand affordance data, using the hand postures of Fig.~\ref{fig:postures}. This structure is similar to the one that gave us the best generalization results in our previous work~\cite{goncalves:2014:icdl}, thanks to its dimensionality reduction which reduced the number of edges, therefore the computational complexity of training and testing, and most importantly it reduces the amount of training data required to observe the emergence of some learning effect.

All the nodes of the network, except for $\EffectX$ and $\EffectY$, are entered as interventional variables, which means that we force a node to take a specific value~(i.e., they are the prior inputs, when reasoning on the effects), thereby effectively severing its incoming arcs. The two effect nodes can be seen as the outputs~(when we perform inference queries about their values).
The inference performed on the networks is of the type $p(\Effect \mid \parents(\Effect))$, which, considering the topology of our network from Fig.~\ref{fig:pca_bayesian_network}, amounts to this marginalization over our two effect nodes (horizontal and vertical displacement of the object):
\begin{equation} \label{eq:effect_query}
    p(\EffectX, \EffectY \mid M, O, A),
\end{equation}
where $M$~is the vector of features of the manipulator~(if the manipulator is a hand we refer to its feature vector as~$H$, if it is a tool we refer to its feature vector as~$T$), $O$~is the vector of features of the target object, $A$~is the motor action identifier.

\section{Results}
\label{sec:results}

In this section, we present the results obtained from our hand affordance network which is trained with robot experience on objects using different hand postures and motor actions, and we assess its performance.

\subsection{Dataset Inspection}

Our experimental data is obtained by making manipulation experiments on an iCub humanoid robot, in a setup like the one shown in Fig.~\ref{fig:cover}, using its left arm for data collection. We consider 4~motor actions A (tapFromRight, tapFromLeft, draw, push), 2~objects O (lego piece, pear), 3~hand postures H (straight fingers, bent fingers, arched fingers). We extract the visual features from both O and H (before performing the actions). The dataset is publicly available at~\url{http://vislab.isr.tecnico.ulisboa.pt/} $\rightarrow$ Datasets.

\begin{figure*}
    \subfloat[][Motion caused with the robot \emph{hands} when using different actions and hand postures, as observed when interacting with 2~objects multiple times in the new experiments from this paper.]
    { \includegraphics[width=0.45\linewidth]{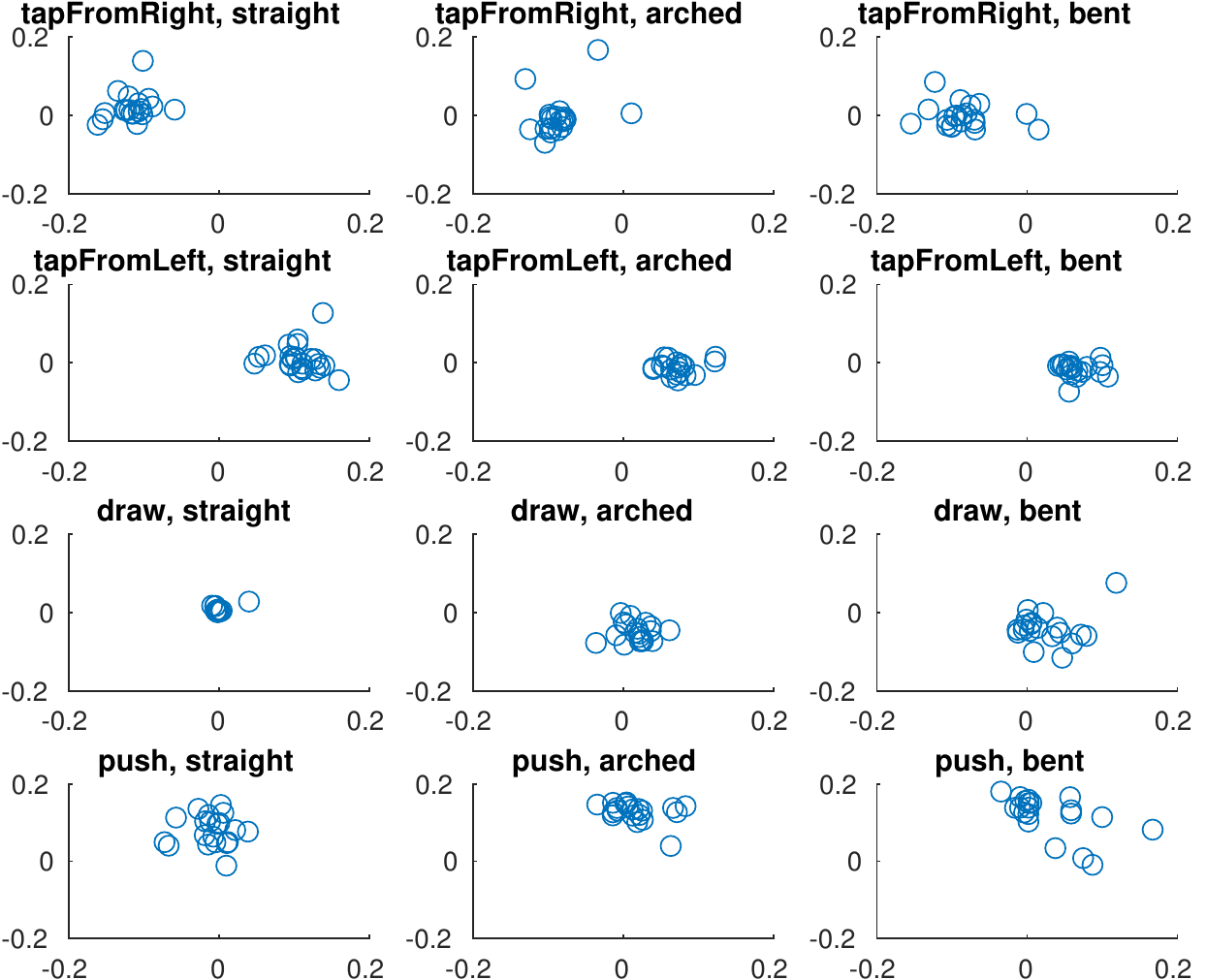} \label{fig:effect_data:hands} } \quad
    \subfloat[][Motion caused with \emph{tools} when using different actions and tool types, from~\cite{dehban:2016:eccvws}. Here we show only the interactions with 2~objects, to be consistent with Fig.~\ref{fig:effect_data:hands}.]
    { \includegraphics[width=0.45\linewidth]{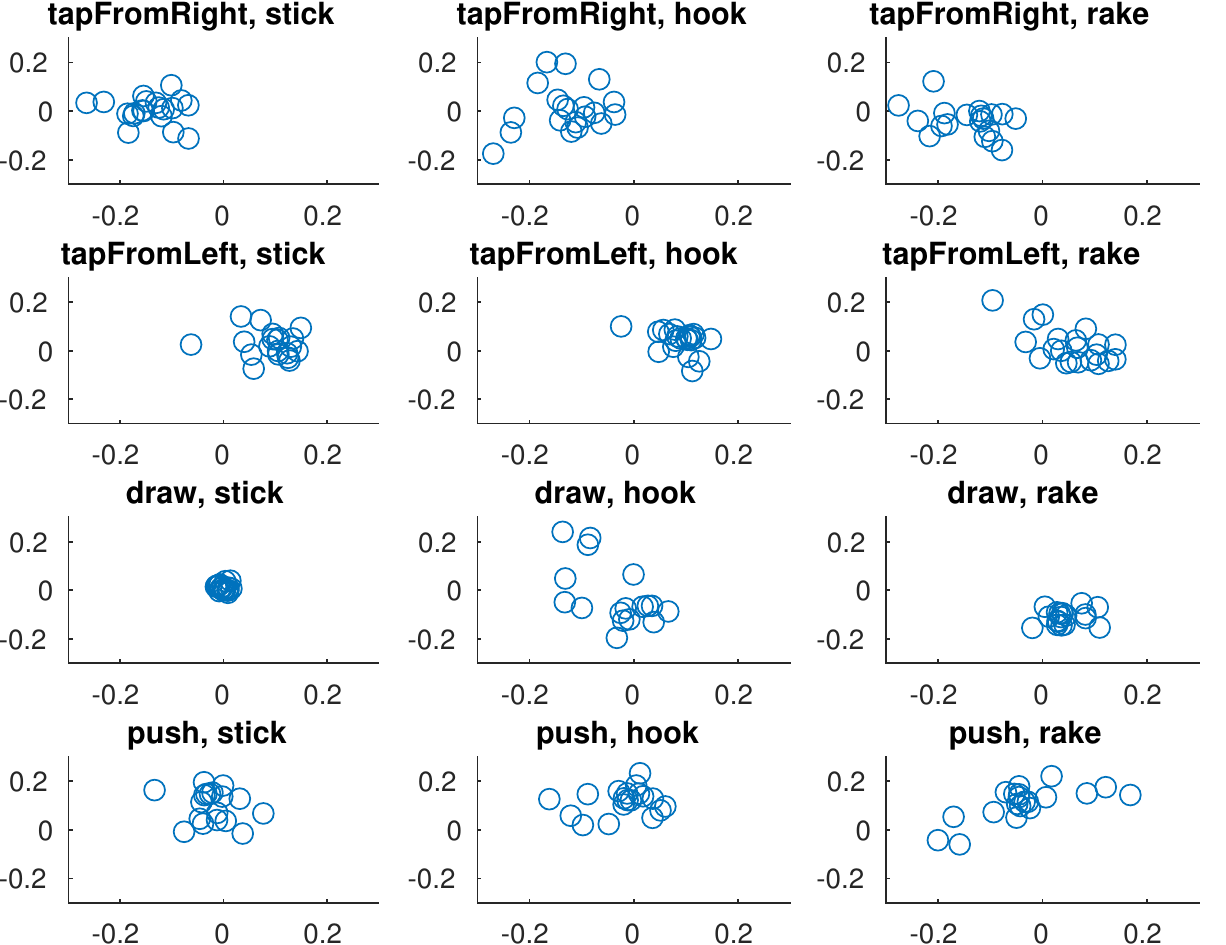} \label{fig:effect_data:tools} }
    \caption{Motion caused by different robotic manipulators when using different actions and manipulator morphologies: in Fig.~\ref{fig:effect_data:hands} we use different hand postures, whereas in Fig.~\ref{fig:effect_data:tools} we vary tool types for comparison. Each plot displays the geometrical displacement along horizontal and vertical direction~(in meters, measured from the object initial position) from the point of view of the robot~(the robot is at the~0 in the x-axis marker). For example, tapping an object from the right~(tapFromRight action) usually results in making the object shift to the left direction; drawing~(i.e., pulling) an object closer only works if the manipulator morphology is appropriate.}
    \label{fig:effect_data}
\end{figure*}

\begin{figure}
    \centering
    \includegraphics[width=0.9\linewidth]{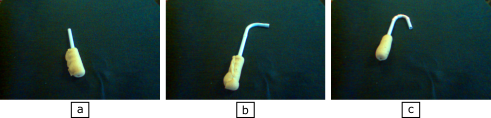}
    \caption{The three tools used in~\cite{dehban:2016:eccvws}:
    (a)~stick, (b)~rake and (c)~hook. They provide different affordances when grasped by the hand of the robot and employed for performing motor actions onto objects. In this paper, we consider these tool affordances for comparison with our novel hand affordances.}
    \label{fig:tools}
\end{figure}

In Fig.~\ref{fig:effect_data} we show the distributions of the motion effects onto target objects caused by the robot influence when it touches objects with its manipulator. In particular, Fig.~\ref{fig:effect_data:hands} shows the effects of using the different hand postures. For comparison, Fig.~\ref{fig:effect_data:tools} depicts the effect of using the elongated tools (Fig.~\ref{fig:tools}) on the same objects. Visual inspection reveals the similarities in the effect of using tools or hands, for example, tapping from left usually results in the object moving to the right. Another prominent similarity is that drawing with a stick or with straight hand posture causes only minimal movement.

\subsection{Data Augmentation}

Similar to previous research~\cite{goncalves:2014:icarsc,goncalves:2014:icdl,dehban:2016:eccvws}, we assume that the affordance of an object and of a robot manipulator is viewpoint-invariant. By exploiting this notion, it is possible to artificially augment the trials data using \emph{multiple views} of manipulators and objects. In all of the following experiments, we have used at least 10~viewpoints of each object and manipulator, effectively multiplying the number of available samples by more than~100 times.

\subsection{Effect Prediction}
\label{sec:results:effect_prediction}

One way to assess the quality of the learned Bayesian Network of Fig.~\ref{fig:pca_bayesian_network} is to predict the effect distribution, given the descriptors of manipulator, object, and action, i.e., the direct application of~\eqref{eq:effect_query}. To this end, we have empirically divided the effect distribution along each axis into five bins~(a list of the hyper-parameters that we used for training our network is reported in Table~\ref{tab:pca_hyperparams} for reproducibility). If the network predicts the correct effect bin out of five, it gets a point. Since there exist two axis directions, a random predicting machine would be correct~1/25 of the time.

\begin{table*}
    \centering
    \caption{Hyper-parameters used to train the Bayesian Network of Fig.~\ref{fig:pca_bayesian_network} for predicting the distribution of the effects.}
    \label{tab:pca_hyperparams}
    \begin{tabular}{*{2}{l}} 
    \toprule
    parameter & value (and comment) \\
    \midrule
    number of Principal Component Analysis~(PCA) blocks & one for manipulator, one for object \\
    number of components of each PCA block & $2$ \\
    number of discretization values (bins) of each PCA component & $2$ \\
    number of discretization values (bins) of each $\Effect$ node & $5$ \\
    intervals (in meters) of the $\Effect$ bins & $\interval[open left]{-\infty}{-0.06}$, $\interval[open left]{-0.06}{-0.025}$, $\interval[open left]{-0.025}{0.025}$, $\interval[open left]{0.025}{0.06}$, $\interval[open]{0.06}{\infty}$\\
    \bottomrule
    \end{tabular}
\end{table*}

Using the same network parameters but training with different data, the following accuracy scores were obtained:

\begin{itemize}
    \item train 80\% hand, test 20\% hand: accuracy 72\%;
    \item train 80\% tool, test 20\% tool: accuracy 58\%;
    \item train 100\% hand, test 100\% tool: accuracy 53\%.
\end{itemize}

To explain these scores, we note that motor control on the iCub is noisy, and actions on this platform are not deterministic or repeatable~(e.g., when commanding the robot twice starting from an initial position, the same motor command can produce two slightly different configurations).
Even so, in the accuracies and in Fig.~\ref{fig:effect_data} we see that tool effects are more varied than hand effects, making tools less reliable~(i.e., more noisy) than hands. Nevertheless, by only training on the hand data, we obtain an accuracy that is comparable with the case where the network is trained on tool data, demonstrating the generalization of our proposed method.

\subsection{Tool Selection}

One question that this paper tries to embark upon is the following: if an agent gains the knowledge of how her hand postures can affect the environment, can she generalize this knowledge to other tools which look similar to her hands? To answer this question in the scope of the presented scenario, we conduct the following experiment. We suppose that an agent has defined a goal, for example to pull an object towards herself. She knows that the correct action for this task will be to \emph{draw}, however the object is out of hand's reach and one of the presented tools~(see Fig.~\ref{fig:tools}) must be selected for the task.

In this scenario, an agent looks at the available tools and the target object and performs a mental simulation of the known action. A tool is selected if it is expected to cause a movement of the target object \emph{along the desired direction}, and it is rejected if no movement is predicted, or if the object is predicted to move against the desired direction. Because in this work we divide the direction into five bins~(see Sec.~\ref{sec:results:effect_prediction}), we compare the sum of the predictions in the two desired-movement bins against the sum of the predictions in the remaining bins. Since there was no interaction with the tool, it is necessary to generalize from the knowledge of previous hand explorations to tools in a zero-shot manner.

As an example of a successful generalization, the agent should predict that drawing an object with a stick is pointless because she has already experimented drawing objects with a straight hand posture, and the visual descriptors of straight hands are similar to those of a stick. Table~\ref{tab:tool_selection_results} shows the result of this inquiry. We have also implemented a baseline in which the agent has already experienced with the tools and is asked to select the correct tool. As expected, all the tools can be used for the desired effects, and it is only the draw action which requires a tool with a specific shape. The numbers are normalized, as they correspond to different views of the tool and object, and they reflect the percentage of the cases where that specific tool was selected.

\begin{table}
    \centering
    \caption{Tool selection results obtained from our ``hand to tool''~(HT) network, compared to ones obtained from the baseline ``tool to tool''~(TT) network~\cite{dehban:2016:eccvws}.}
    \label{tab:tool_selection_results}
    \begin{tabular}{*{4}{l}} 
    \toprule
    action       & stick             & hook           & rake \\
    \midrule
    tapFromRight & HT: $1.0$         & HT: $1.0$         & HT: $1.0$ \\
                 & (TT: $1.0$)       & (TT: $1.0$)       & (TT: $1.0$) \\
                 &                   &                   & \\
    tapFromLeft  & HT: $1.0$         & HT: $1.0$         & HT: $1.0$ \\
                 & (TT: $1.0$)       & (TT: $1.0$)       & (TT: $1.0$) \\
                 &                   &                   & \\
    draw         & HT: $\bm{0.5385}$ & HT: $\bm{0.6154}$ & HT: $\bm{1.0}$ \\
                 & (TT: $0.1538$)    & (TT: $0.1538$)    & (TT: $0.4615$) \\
                 &                   &                   & \\
    push         & HT: $1.0$         & HT: $1.0$         & HT: $1.0$ \\
                 & (TT: $1.0$)       & (TT: $1.0$)       & (TT: $1.0$) \\
    \bottomrule
    \end{tabular}
\end{table}

\section{Conclusions}
\label{sec:conclusions}

We introduce a computational model of hand affordances: a Bayesian Network that relates robot actions, visual features of hand postures, visual features of objects and produced effects, allowing a humanoid robot to predict the effects of different manual actions. Interestingly, we show how this knowledge, acquired by the robot through autonomous exploration of different actions, hand postures and objects, can be generalized to tool use, and employed to estimate the most appropriate tool to obtain a desired effects on an object, among a set of tools that were never seen before.

A few important comments have to be made to better focus the scope of our contribution. Our results show that, in some specific cases, it is indeed possible to generalize what was learned about hand affordances to tools that were never seen before. Clearly, this is limited to a subset of all the possible human-made tools that a humanoid robot could see and possibly use; however, the previous knowledge about hand affordances can give the robot the possibility to make a good initial estimate of how a tool could be used. It would be very interesting to investigate how further sensorimotor experience with tools can be integrated in the learned model, and possibly permit better predictions. Also, it may be interesting to explore the opposite direction: can the knowledge acquired with a specific tool be re-used to estimate the effects of manual actions without the tool, or to shape the robot hand in the best posture to achieve some effects~(i.e., from tool to hand affordances)? 

As we made clear in the introduction, to the best of our knowledge, there is no clear experimental evidence that human children exploit visual information about their hand postures to estimate the possible affordances of an external object~(i.e., a tool); neither is there evidence that visual information extracted from successfully used tools is employed to control hand postures. Also, although the early manipulation attempts of the child are probably a necessary prerequisite for effective tool use, a great deal of knowledge is indeed acquired during direct sensorimotor exploration using the tools, and during observations of tool actions performed by others~(e.g., by the parents). However, our results can serve a twofold purpose for the developmental robotics community: (i)~we propose a robot learning framework that presents practical advantages for robot autonomy, at least in the limited number of situations we experimented, since it permits to generate meaningful predictions about a non-finite set~(i.e., tools) from experiences in a finite set~(i.e., hand postures); and (ii)~we lay the basis for future discussion and research in developmental psychology and cognitive science about the possible existence of similar phenomena in humans.

\section*{Acknowledgments}

This work was partially supported by the FCT project UID/EEA/50009/2013 and grant PD/BD/135115/2017. We thank the anonymous reviewers for their helpful suggestions.

\printbibliography

\end{document}